\documentclass[10pt,twocolumn,letterpaper]{article}
\usepackage[accsupp]{axessibility}
\usepackage[pagenumbers]{iccv}
\usepackage{natbib}
\usepackage{xcolor} 
\usepackage{pst-text}
\usepackage{amssymb}
\usepackage{pifont}
\usepackage{colortbl}
\usepackage{multirow}
\usepackage{mathrsfs}
\usepackage{titletoc}
\usepackage{placeins}
\usepackage[most]{tcolorbox}
\newcommand{\pub}[1]{\color{gray}{\scriptsize{[{#1}]}}}
\usepackage[misc]{ifsym} 
\newcommand{\cmark}{\ding{51}\xspace}%
\newcommand{\xmarkg}{\textcolor{lightgray}{\ding{55}}\xspace}%
\definecolor{myorange}{RGB}{255,100,3}
\definecolor{mygray}{gray}{.85}
\definecolor{mygray1}{gray}{.7}
\definecolor{mygray2}{gray}{.93}
\definecolor{mygray3}{gray}{.90}

\definecolor{tablecolor}{rgb}{.9,.9,1}

\newcommand{\ourdataset}{\texttt{MOVE}\xspace}
\newcommand{\ourmodel}{DMA\xspace}

\newcommand{\videonum}{4,300\xspace}
\newcommand{\framenum}{261,920\xspace}
\newcommand{\objectnum}{5,135\xspace}
\newcommand{\masknum}{314,619\xspace}
\newcommand{\motionnum}{224\xspace}
\newcommand{\objcategorynum}{88\xspace}

\newcommand{\metricj}{$\mathcal{J}$\xspace}
\newcommand{\metricf}{$\mathcal{F}$\xspace}
\newcommand{\metricjf}{$\mathcal{J}\&\mathcal{F}$\xspace}

\newcommand{\myrule}{\specialrule{0.08em}{.05em}{.05em}}
\newcommand{\mymidrule}{\specialrule{0.03em}{.05em}{.05em}}

\newcommand{\myparagraph}[1]{{\vspace{.1em} \noindent \bf #1}}

\definecolor{cvprblue}{rgb}{0.21,0.49,0.74}
\usepackage[pagebackref,breaklinks,colorlinks,allcolors=cvprblue]{hyperref}

\def\paperID{xxxx} 
\def\confName{ICCV}
\def\confYear{2025}

\begin{document}
\title{\ourdataset: Motion-Guided Few-Shot Video Object Segmentation}

\author{
Kaining Ying\footnotemark[1]
\quad
Hengrui Hu\footnotemark[1]
\quad
Henghui Ding~\!$^{\textrm{\Letter}}$
\vspace{.6mm}
\\
Fudan University, China
\vspace{.6mm}
\\
\href{https://henghuiding.com/MOVE/}{https://henghuiding.com/MOVE/}
}

\twocolumn[{
\renewcommand\twocolumn[1][]{#1}%
\maketitle
\begin{center}
    \centering
    \captionsetup{type=figure}
    \includegraphics[width=0.999\textwidth]{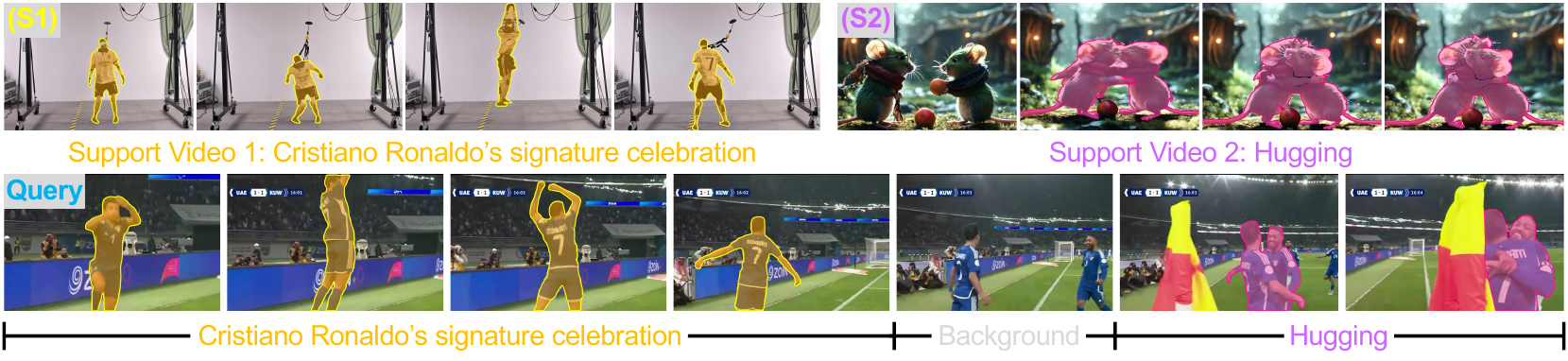}
    \vspace{-5mm}
    \captionof{figure}{\small 
    We propose a new benchmark for \textbf{MO}tion-guided Few-shot \textbf{V}ideo object s\textbf{E}gmentation~(\textbf{\ourdataset}).
    In this example, given two support videos showing distinct motion patterns (S1: Cristiano Ronaldo's signature celebration~\cite{asap}, S2: hugging), our benchmark aims to segment target objects in the query video that perform the same motions as in the support videos. 
    \textbf{\ourdataset} provides a platform for advancing few-shot video analysis and perception by enabling the segmentation of diverse objects that exhibit the same motions.
    }
    \label{fig:teaser}
    \vspace{2.96mm}
\end{center}
}]

\renewcommand{\thefootnote}{\fnsymbol{footnote}}
\footnotetext[1]{Equal contribution.}
\footnotetext[0]{${\textrm{\Letter}}$ Henghui Ding (henghui.ding@gmail.com) is the corresponding author with
the Institute of Big Data, College of Computer Science and Artificial Intelligence, Fudan University, Shanghai, China.}

\begin{abstract}
This work addresses motion-guided few-shot video object segmentation~(FSVOS), which aims to segment dynamic objects in videos based on a few annotated examples with the same motion patterns. Existing FSVOS datasets and methods typically focus on object categories, which are static attributes that ignore the rich temporal dynamics in videos, limiting their application in scenarios requiring motion understanding. To fill this gap, we introduce \textbf{\ourdataset}, a large-scale dataset specifically designed for motion-guided FSVOS. Based on \ourdataset, we comprehensively evaluate 6 state-of-the-art methods from 3 different related tasks across 2 experimental settings. Our results reveal that current methods struggle to address motion-guided FSVOS, prompting us to analyze the associated challenges and propose a baseline method, Decoupled Motion-Appearance Network~(\textbf{\ourmodel}). Experiments demonstrate that our approach achieves superior performance in few-shot motion understanding, establishing a solid foundation for future research in this direction. 
\end{abstract}

\section{Introduction}
\label{sec:intro}
Few-shot video object segmentation (FSVOS)~\cite{fsvos,mao2024fewshot,luo2024exploring,liu2023multigrained,siam2022temporal} aims to segment objects of unseen classes in videos using only a few annotated examples. FSVOS is a relatively underexplored yet promising field.
By requiring only minimal supervision, FSVOS significantly reduces the need for extensive labeled datasets while enabling rapid adaptation to new object classes.
With these advantages, it shows great potential in autonomous driving, robotics, surveillance, augmented reality, and media production~\cite{zhou2022survey}.

Previous FSVOS methods~\cite{fsvos,hpan,tti} are semantic-centric and primarily focus on object categories, associating query videos with support sets based on object class. For example, given support images containing pandas, these methods aim to segment all pandas in the query video regardless of their individual characteristics. 
This semantic-centric paradigm, similar to the widely-studied few-shot image segmentation~(FSS)~\cite{aenet,panet,cyctr,sccan}, largely overlooks the crucial temporal dynamics inherent in videos, such as object motions and temporal dependencies, thus limiting the advancement of FSVOS research.

\begin{figure}
    \centering
    \includegraphics[width=0.998\linewidth]{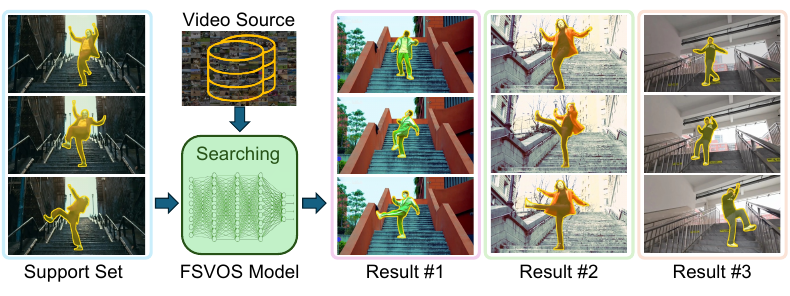}
    \vspace{-8mm}
    \caption{
    Given a motion of interest, our approach enables retrieval and indexing of relevant videos and their corresponding objects from the internet or personal collections. Notably, these motions of interest can be novel actions that are difficult to describe accurately using single-frame images or text alone.
    }
    \label{fig:fig_intro}
    \vspace{-1.6mm}
\end{figure}

We emphasize the fundamental role of motion patterns in videos, which cannot be adequately captured by static image-based segmentation approaches. 
Consider Cristiano Ronaldo's celebration motion shown in~\Cref{fig:teaser}~(S1), such dynamic patterns can only be properly represented through video sequences.~Unlike previous few-shot video segmentation methods that focus on object categories, \eg, \textit{robot} or \textit{mouse} in the support videos of \Cref{fig:teaser}, we explore to segment objects based on their motion patterns, \eg, \textit{hugging}, regardless of their object categories.~This enables us to recognize and segment objects performing the same motions, which facilitates novel motion-based object-level retrieval in video, as shown in \Cref{fig:fig_intro}.

Recent referring video object segmentation (RVOS)~\cite{mevis,dshmp,motion_cao,mutr,OmniAVS,ReferringSurvey,primitivenet} methods explore motion-guided expressions to segment target objects in videos. However, these methods face inherent limitations when dealing with novel or complex motions that are difficult to describe textually. In contrast, such motions can be effectively characterized by providing a reference video example, such as the distinctive dance sequence from the movie in \Cref{fig:fig_intro}. While widely recognized actions eventually receive distinctive names, \eg, ``\textit{CR7's celebration}'' and ``\textit{Joker's dance}'', this does not contradict our approach.~A video is worth a thousand words, particularly for newly emerging actions that have not yet gained widespread recognition. 

In light of this, we propose \textbf{\ourdataset}, a new large-scale motion-guided FSVOS dataset containing \motionnum motion categories, \videonum videos with \framenum frames in total, and \masknum high-quality segmentation masks annotating \objectnum objects across \objcategorynum object categories.
This dataset is designed to capture diverse motion patterns, facilitating the development and evaluation of motion-centric FSVOS methods.

By adapting existing methods~\cite{fsvos,hpan,mevis} to \ourdataset, we find that \ourdataset presents great challenges in understanding and matching motions between support and query videos.
Understanding motions in support videos requires comprehensive analysis of the entire video sequence, rather than relying on static single-frame semantic recognition. 
Furthermore, effectively extracting motion-related prototypes presents another significant challenge, as existing methods primarily focus on extracting semantic features while overlooking the dynamic information inherent in support videos.
To address these challenges, we propose a Decoupled Motion-Appearance~(\ourmodel) module for extracting temporally decoupled motion-appearance prototypes, enabling the model to focus more on object motions rather than object appearance.~Experiments demonstrate that our proposed \ourmodel helps the model learn motion-centric features, thereby effectively improving model performance. We conduct a comprehensive evaluation of 6 state-of-the-art methods from 3 different related tasks across 2 experimental settings in 2 backbones, demonstrating the superiority of the proposed \ourmodel in few-shot motion understanding.

In summary, this work makes the following three main contributions:
\textbf{i)} We introduce \ourdataset, a motion-guided few-shot video object segmentation dataset that shifts the focus from static object categories to dynamic motion understanding.
\textbf{ii)} We propose \ourmodel, a method based on decoupled motion and appearance, which demonstrates effective few-shot motion understanding and achieves strong performance on the proposed \ourdataset.
\textbf{iii)} We conduct comprehensive experiments, benchmarking 6 baselines on \ourdataset, providing a solid foundation for future research.

\section{Related Work}
\label{sec:related_work}

\subsection{Video Object Segmentation}
\begin{table*}[t!]
  \centering
  \small
  \setlength\tabcolsep{4.7pt}
  \renewcommand\arraystretch{1.16}
  \caption{Comparison of related few-shot video object segmentation/detection datasets~\cite{fsvod,tti,fsvos} with our proposed dataset \ourdataset.}
  \label{tab:rel-dataset}
  \vspace{-3mm}
  \begin{tabular}{lccccccccc}
  \myrule
      \textbf{Dataset} & \textbf{Venue} & \textbf{Label Type} & \textbf{Annotation} & \textbf{Support Type} & \textbf{Categories} & \textbf{Videos} & \textbf{Objects} & \textbf{Frames} & \textbf{Masks} \\ 
      \hline
      FSVOD-500~\cite{fsvod} & \pub{ECCV'22} & Object & Box & Image & \textbf{500} & 4,272 & 4,663 & \ \ 96,609 & 104,495\\
      YouTube-VIS~\cite{YouTubeViS,fsvos} & \pub{CVPR'21} & Object & Mask & Image & \ \ 40 & 2,238 & 3,774 & \ \ 61,845 & \ \ 97,110 \\   
      MiniVSPW~\cite{tti} &  \pub{IJCV'25} & Object & Mask & Image & \ \ 20 & 2,471 & - & \textbf{541,007} & - \\ 
      \hline
      \rowcolor{tablecolor} \textbf{\ourdataset (ours)}  & \pub{ICCV'25} & Motion & Mask & Video & \motionnum & \textbf{\videonum} & \textbf{\objectnum} & \framenum & \textbf{\masknum} \\  
  \myrule
  \end{tabular}
  \vspace{-3mm}
\end{table*}

Video Object Segmentation (VOS)~\cite{mose,MOSEv2,davis,lvosv1,lvosv2} aims to track and segment the corresponding objects in a video sequence, given the object mask in the initial frame.
Early deep neural network (DNN)-based methods, such as OSVOS~\cite{caelles2017one} and MoNet~\cite{xiao2018monet}, fine-tuned network parameters during inference to model inter-frame correlations. Methods like OSMN~\cite{yang2018efficient} and LML~\cite{bhat2020learning} used the first frame with a mask as a prompt to generate a prototype for pixel-level matching with subsequent frames.
Recent trends have shifted toward memory-based methods. STM~\cite{oh2019video} introduced memory modules to store historical frame information, while STCN~\cite{cheng2021rethinking} enhanced memory usage efficiency.~XMem~\cite{cheng2022xmem} and Cutie~\cite{cheng2024putting} further improved memory mechanisms with multiple granularities and object-specific storage.~Recently, SAM2~\cite{ravi2024sam2} emerged as a large video model extending SAM~\cite{kirillov2023segment}, achieving significant performance improvements.
While previous VOS methods have achieved milestone progress, their feasibility and scalability are still constrained by the need for large volumes of densely annotated masks. Additionally, many methods struggle with out-of-domain inputs. In contrast to conventional VOS settings, we focus on few-shot settings to significantly reduce annotation costs and improve generalization to a wider range of scenarios.

\subsection{Few-Shot Video Object Segmentation}
Few-shot video object segmentation (FSVOS)~\cite{fsvos,mao2024fewshot,luo2024exploring,liu2023multigrained,siam2022temporal} has emerged as a promising solution to address the heavy dependency on pixel-wise annotations in traditional VOS. Given an annotated support set, FSVOS aims to segment novel object categories unseen during training in query videos with only a few prompt images and masks. Recent FSVOS methods mainly focus on prototype learning~\cite{fsvos,hpan,liu2023multigrained} or affinity calculation~\cite{siam2022temporal}. DANet~\cite{fsvos} first defined FSVOS and proposed sampled query agents for attention. TTI~\cite{siam2022temporal} and VIPMT~\cite{liu2023multigrained} focused on temporal consistency through prototypes at different granularities. HPAN~\cite{tang2023holistic} and CoCoNet~\cite{luo2024exploring} further improved temporal modeling with graph attention and optimal transport respectively.
While these studies advance few-shot segmentation for novel object categories, they are inherently category-centric, limiting their real-world applicability. In contrast, we propose a motion-centric approach that prioritizes the motion over the object category. In our new task, \ourdataset, the support set consists of videos and masks specifying a particular motion, and the model segments objects performing the same motion in query videos regardless of their categories, enabling generalization to both novel motions and objects.

\subsection{Motion-centric Tasks}
Motion understanding has evolved as a core research direction in video analysis, progressing from early human-centered action recognition~\cite{scovanner20073, wang2013action, zhu2016key} to more complex tasks including action detection~\cite{shou2016temporal, zhao2021video, yang2020revisiting}, spatio-temporal localization~\cite{kopuklu2019you, pan2021actor, chen2023efficient}. 
Recent LVLMs~\cite{motionbench,mmtbench,convbench} also pay attention to temporal motion-related tasks.
However, these methods require extensive training data and cannot accurately segment target objects in videos. Recently, referring video object segmentation~\cite{mevis,dshmp,motion_cao,sola} has explored using motion-related expressions to segment target objects, but expressions often fail to accurately describe novel motions. Our proposed \ourmodel can learn novel actions with minimal data to segment target objects in query videos, enabling broader applications across diverse real-world scenarios.

\section{\ourdataset Benchmark}
\label{sec:dataset}

\subsection{Task Setting}
\myparagraph{Revisit of FSVOS.}
Few-Shot Video Object Segmentation (FSVOS) \cite{fsvos,hpan,tti} aims to learn a segmentation model that can generalize to novel object categories with limited labeled examples. 
The framework operates on two disjoint data splits: a base class training set $\mathcal{D}_{train}$ and a novel class test set $\mathcal{D}_{test}$. The evaluation protocol involves episodes, where each episode comprises a support set $\mathcal{S}$ and a query set $\mathcal{Q}$. Specifically, the support set encompasses $K$ pairs of images and their corresponding masks $\{(I^s_k, M^s_{k,c})\}_{k=1}^K$ extracted from separate videos, with $I^s_k$ denoting the $k$-th support image and $M^s_{k,c}$ representing its associated mask for class $c$. The query set contains a video with $T$ frames $\{(I^q_t, M^q_{t,c})\}_{t=1}^T$, where $I^q_t$ indicates the $t$-th frame and $M^q_{t,c}$ denotes its ground truth mask. The objective of FSVOS is to leverage the visual and semantic cues from support samples to accurately segment target objects in query video frames.

\myparagraph{Extension.} 
The proposed \ourdataset focuses on motion categories rather than object categories.~Since static images inherently lack the capacity to represent temporal dynamics, we extend the support set $\mathcal{S}$ to contain $K$ video-mask pairs $\{(V^s_k, M^s_{k,c})\}_{k=1}^K$ sampled from different videos, where each video clip $V^s_k$ demonstrates a motion pattern with corresponding mask sequence $M^s_{k,c}$ of motion-class $c$. The query set $\mathcal{Q}$ remains as a video sequence of $T$ frames $\{(I^q_t, M^q_{t,c})\}_{t=1}^T$. This formulation shifts the focus from {static appearances} to {temporal modeling}, emphasizing motion as the core feature for video understanding.

\subsection{Dataset Annotation}
\label{sec:dataset_collection}
\myparagraph{Vocabulary Collection.}
Following previous video recognition datasets~\cite{haa500,kinetics,ucf101}, 
we build a hierarchical vocabulary set with four areas: \textit{daily actions}, \textit{sports}, \textit{entertainment activities}, and \textit{special actions}. Each category follows three criteria: fine-grained, mutual exclusion (clear semantic boundaries), and novelty (not well covered in existing datasets). This systematic classification lays the foundation for motion-guided few-shot video segmentation tasks. 

\myparagraph{Video Clip Collection.}
Videos in \ourdataset are sourced from two parts: 
\textbf{i)} public action recognition datasets~\cite{haa500, li2021multisports, parikh2024idd, li2021ai, jian2020visual, zhang2018egogesture, wlasl, chen2023mammalnet, kuehne2011hmdb, bit} and
\textbf{ii)} internet videos under a \textit{Creative Commons License}.
During the selection process, we followed these criteria: videos should have clear motion boundaries, diverse scenes, and varied subject categories.

\myparagraph{Mask Annotation.}
For videos without preexisting masks, we recruited well-trained annotators to label high-quality masks with the assistance of a state-of-the-art VOS segmentation model \cite{sam2} on an interactive annotation platform. 

\subsection{Data Statistics and Analysis}
\label{sec:data_analysis}
As shown in \Cref{tab:rel-dataset}, our \ourdataset benchmark contains 224 action categories across four domains (daily actions, sports, entertainment activities, and special actions), with \videonum video clips, \objectnum moving objects, \framenum frames, and \masknum mask annotations. Compared to existing object-centric datasets, \ourdataset features video-level support samples and motion-based categories, while maintaining comparable scale in terms of videos and annotations. Each video clip is equipped with high-quality pixel-level mask annotations, capturing diverse scenes, subjects (person, vehicle, animal, \etc), and motion complexities. 
For more statistics, please refer to the supplementary materials.

\section{Methodology}
\label{sec:dma}

\begin{figure}[t!]
    \centering
    \includegraphics[width=0.98\linewidth]{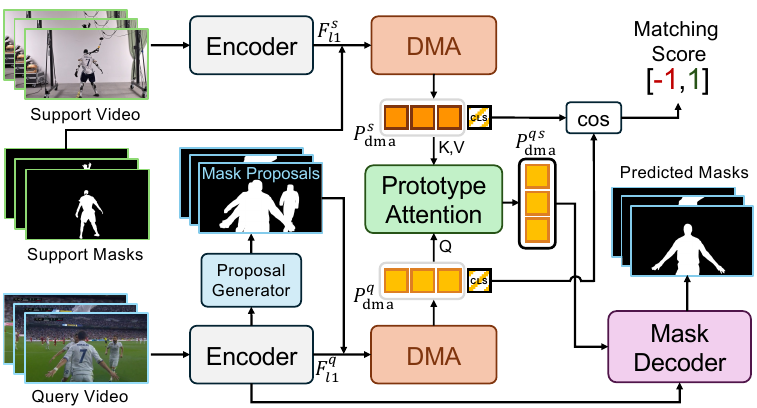}
    \vspace{-1.6mm}
    \caption{Overview of our proposed method.}
    \label{fig:main_fig}
    \vspace{-3mm}
\end{figure}

\subsection{Overview}
\label{sec:overview}
As shown in \Cref{fig:main_fig}, the proposed method consists of five main components: 1) a shared encoder for extracting multi-scale features from both support and query video frames, 2) proposal generator for obtaining coarse mask proposals of the query video, 3) a shared DMA module for extracting decoupled motion-appearance prototypes, 4) prototype attention module for facilitating interaction between support and query prototypes, and 5) mask decoder for generating the final segmentation masks of the query video. In the following sections, we describe each component in detail.
For simplicity, we describe our method in the \textit{1-way-1-shot} setting, although it can be easily extended to \textit{N-way-K-shot} scenarios. 
Given a support video clip with $T_s$ frames $\{I^s_t\}_{t=1}^{T_s}$ and corresponding mask sequence $\{M^s_t\}_{t=1}^{T_s}$, along with a query video clip containing $T_q$ frames $\{I^q_t\}_{t=1}^{T_q}$, 
our goal is to segment out the target object mask sequence $\{\hat{M}^q_t\}_{t=1}^{T_q}$ in the query video that exhibits the same motion pattern as the object in the support video.

\subsection{Encoder and Proposal Generator}
\myparagraph{Encoder.}~Our encoder $\mathcal{E}$ combines a backbone~\cite{resnet,videoswin} with a feature pyramid network~\cite{fpn} to extract multi-scale features from both the support and query videos as follows:
\begin{equation}
F_{l1,t}, F_{l2,t}, F_{l3,t}, F_{l4,t} = \mathcal{E}(I_t), \ \ t=1,\dots,T,
\label{eq:encoder}
\end{equation}
where $F_{li,t}$ is the $i$-th layer feature of the $t$-th frame $I_t$. $T$ denotes the total number of frames. $F_{l1,t}$, $F_{l2,t}$, $F_{l3,t}$, and $F_{l4,t}$ correspond to features at resolutions of $1/4$, $1/8$, $1/16$, and $1/32$ of input resolution, respectively.

\myparagraph{Proposal Generator.}
This module processes multi-scale query features $\{F_{l1}^q, F_{l2}^q, F_{l3}^q, F_{l4}^q\}$ to generate coarse mask proposals. It employs three convolutional blocks at different scales ($1/32$, $1/16$, and $1/8$ resolution) with residual connections. Features are progressively refined through upsampling and fusion, with final predictions generated by a lightweight convolutional head that outputs single-channel proposals. This approach effectively balances multi-scale information utilization and computational efficiency.

\begin{figure}[t!]
    \centering
    \includegraphics[width=0.98\linewidth]{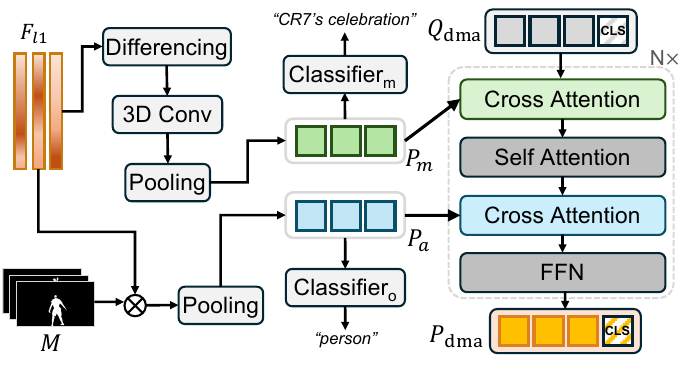}
    \vspace{-1.16mm}
    \caption{Decoupled Motion-Appearance (DMA) Module.}
    \label{fig:dma}
    \vspace{-1.96mm}
\end{figure}

\subsection{Decoupled Motion-Appearance Module}
\label{sec:dma_module}
As shown in \Cref{fig:dma}, DMA module extracts decoupled motion-appearance prototypes for both query and support branches. The module takes the $1/4$ resolution features $F_{l1}$ and corresponding object masks $M$ as input, where the support branch utilizes pre-annotated support masks while the query branch leverages mask proposals generated by the proposal generator.

\myparagraph{Appearance Prototype.}~\ourmodel first extracts appearance prototypes $P_{a}$ by applying mask pooling on feature $F_{l1}$:
\begin{equation}
P_{a} = \frac{\sum_{h,w} F_{l1} \odot M}{\sum_{h,w} M} \in \mathbb{R}^{T \times d},
\label{eq:appearance_proto}
\end{equation}
where $\odot$ denotes element-wise multiplication.

\myparagraph{Motion Prototype.}
\ourmodel then extracts motion prototypes by calculating temporal differences between adjacent frames features, where the temporal difference at the last time step is padded with zeros:
\begin{equation}
\begin{aligned}
D_{l1,t} &= F_{l1,t+1} - F_{l1,t}, \ t=1,\dots,T-1, \\
P_{m} &= \text{Pooling}(\text{Conv3D}(D_{l1}))\in \mathbb{R}^{T \times d},
\end{aligned}
\label{eq:motion_feat}
\end{equation}
where $\text{Conv3D}$ denotes 3D convolutional layers for temporal feature enhancement, and $\text{Pooling}$ is a spatial pooling operation that aggregates the motion features across the spatial dimensions into motion prototypes $P_{m}$.

To guide the learning of discriminative and complementary motion and appearance prototypes, we introduce two auxiliary classification heads:
\begin{align}
p_o &= \text{Classifier}_o(\text{AvgPool}(P_a)) \in \mathbb{R}^{C_o},\\
p_m &= \text{Classifier}_m(\text{AvgPool}(P_m)) \in \mathbb{R}^{C_m},
\end{align}
where $C_o$ is the number of predefined object categories and $C_m$ is the number of motion categories. These classification tasks explicitly guide $P_a$ to encode object-specific appearance information, while $P_m$ focuses on motion-specific temporal dynamics.~This decoupled supervision ensures that the two prototype branches learn complementary features for appearance and motion, respectively.

The extracted motion prototypes $P_m$ are further refined using a transformer-based architecture. As shown in \Cref{fig:dma}, this architecture processes learnable queries $Q_{\text{dma}}$ and a special [\texttt{CLS}] token through multiple transformer layers. Each transformer layer consists of cross-attention modules attending to motion prototypes $P_m$ and appearance prototypes $P_a$, followed by self-attention modules and feed-forward networks (FFN). This process produces the final decoupled motion-appearance prototypes $P_{\text{dma}}$ along with the [\texttt{CLS}] token used for prototype matching.

\subsection{Prototype Attention and Mask Decoder}

\myparagraph{Prototype Attention.}~To fuse prototype features from both support and query videos,
we introduce a prototype attention module that consists of multiple transformer layers. Given the decoupled motion-appearance prototypes $P_{\text{dma}}^s$ and $P_{\text{dma}}^q$,
this module performs cross-attention where $P_{\text{dma}}^q$ serves as queries while $P_{\text{dma}}^s$ serves as keys and values, followed by self-attention on the enhanced $P_{\text{dma}}^q$ features.
Through multiple transformer layers, this iterative process refines the prototypes, facilitating effective information exchange while preserving their distinctive characteristics.
The enhanced prototypes, denoted as $P_{\text{dma}}^{qs}$, are subsequently used for mask generation in Mask Decoder.

\myparagraph{Mask Decoder.}~The Mask Decoder generates segmentation masks by fusing multi-scale features under the guidance of prototypes $P_{\text{dma}}^{qs}$. It enhances features at different scales via cross-attention with prototypes, enabling the features to focus on motion-centric information. These enhanced features are then progressively fused in a top-down manner. This hierarchical design together with motion prototype guidance ensure that both high-level semantic information and low-level spatial details are effectively leveraged, contributing to the accurate prediction of the final mask.

\myparagraph{Matching Score.}
To determine whether the query instance exhibits the same motion as the support instance, we compute a matching score based on the [\texttt{CLS}] tokens from both branches. The matching score is calculated as:
\begin{equation}
S_{\text{match}} = \cos([\texttt{CLS}]_s, [\texttt{CLS}]_q),
\label{eq:matching_score}
\end{equation}
where $\cos(\cdot,\cdot)$ represents the cosine similarity between the [\texttt{CLS}] tokens from support and query branches. The matching score $S_{\text{match}}$ ranges from -1 to 1, with higher values indicating that the query instance is performing the same motion as the support instance, and lower values suggesting different motions.

\section{Experiment}
\label{sec:experiment}

\myparagraph{Evaluation Metrics.}
Following prior works~\cite{mose,fsvos,hpan}, we use \metricjf to evaluate segmentation quality, with \metricj and \metricf measuring IoU and contour accuracy, respectively. For robustness evaluation, we include query samples with empty foreground and adopt N-Acc and T-Acc metrics~\cite{gres} to measure accuracy on empty and non-empty target samples.

\myparagraph{Dataset Settings.}
\ourdataset contains \motionnum motion categories. 
For cross-validation, we split these into 4 folds with two strategies: 
\textbf{O}verlapping \textbf{S}plit (\textbf{OS}) and \textbf{N}on-overlapping \textbf{S}plit (\textbf{NS}) based on node-level motion distribution. 
Please refer to supplementary materials for more details.
 
\myparagraph{Implementation Details.}
Our backbone uses ResNet50 \cite{resnet} pre-trained on ImageNet~\cite{imagenet} and VideoSwin-Tiny~\cite{videoswin} pre-trained on Kinetics-400~\cite{kinetics}.
Following previous work~\cite{fsvos,hpan,ctvis,isda}, we employ both cross-entropy and IoU losses for mask prediction and proposal generation. Additionally, we use cross-entropy loss for the auxiliary classification head and matching score prediction.
We use a learning rate of 1e-5 with a cosine annealing scheduler for optimization. For our main experiments, we train for 240,000 episodes on 3 folds and test on the remaining fold with 20,000 episodes, using both \textit{2-way-1-shot} and \textit{5-way-1-shot} settings as our primary configurations. Unless otherwise specified, ablation studies are conducted with 150,000 episodes, training on 2 folds and testing on the remaining 2 folds, using the \textit{2-way-1-shot} setting on OS. All the experiments are conducted on 4 NVIDIA RTX A6000 (48GB) GPUs.

\begin{table}[t!]
    \footnotesize
    \centering
    \caption{Necessity study of the proposed \ourdataset benchmark. }
    \vspace{-3mm}
    \setlength\tabcolsep{5.16pt}
    \label{tab:necessity_study}
    \begin{tabular}{lccccc}
        \toprule
        \textbf{Methods} & \textbf{Type} & \textbf{Support} & \textbf{Query} & \textbf{YTVIS}~\cite{fsvos}  & \textbf{\ourdataset} \\
        \midrule
        SCCAN~\cite{sccan}     &  FSS       & Image        & Image        &  62.3                & 40.6   \\
        HPAN~\cite{fsvos}      &  FSVOS     & Image        & Video        &  63.0                & 44.4   \\
        HPAN*                  &  FSVOS     & Video        & Video        &  62.7                & 46.3   \\
        LMPM \cite{mevis}      &  RVOS      & Text         & Video        &  62.5                & 41.8   \\
        \bottomrule
    \end{tabular}
    \vspace{-2.6mm}
\end{table}
\begin{table*}[t!]
    \footnotesize
    \centering
    \setlength\tabcolsep{3.4pt}
    \caption{Main results on \ourdataset benchmark with overlapping split (OS) setting. \textbf{Bold} and \underline{underlined} indicate the largest and second largest values under the same backbone, respectively. VSwin-T indicates VideoSwin-T backbone~\cite{videoswin}.}
    \vspace{-3mm}
    \label{tab:MOVENS}
    \begin{tabular}{lccccccccccccccccc}
    \myrule
    \multirow{2}{*}{\centering Methods} &
      \multirow{2}{*}{\centering Venue} &
      \multirow{2}{*}{\centering Type} &
      \multirow{2}{*}{\centering Backbone} &
      \multicolumn{3}{c}{Mean (2-way-1-shot)} &
      \multicolumn{4}{c}{\metricjf (2-way-1-shot)} &
      \multicolumn{3}{c}{Mean (5-way-1-shot)} &
      \multicolumn{4}{c}{\metricjf (5-way-1-shot)} \\
      \cmidrule(lr){5-7} \cmidrule(lr){8-11} \cmidrule(lr){12-14} \cmidrule(lr){15-18}
                       &         &       &             & \metricjf & T-Acc & N-Acc & 1    & 2    & 3    & 4    & \metricjf & T-Acc & N-Acc & 1    & 2    & 3     & 4    \\
      \midrule
            LMPM~\cite{mevis}       & \pub{ICCV'23} & RVOS  & ResNet50   & 41.8 & 93.1 & \ \ 5.3 & 45.2 & 42.1 & 40.7 & 39.1 & 26.3 & 98.3 &\ \  2.6 & 27.5 & 31.7 & 22.7 & 23.3 \\
            CyCTR~\cite{cyctr}      & \pub{ECCV'24} & FSS   & ResNet50   & 34.4 & \underline{98.4} & \ \ 1.2 & 32.8 & 34.4 & 35.7 & 34.5 & 22.5 & \underline{99.2} &\ \  0.1 & 23.1 & 21.5 & 20.8 & 24.7 \\
            SCCAN~\cite{sccan}      & \pub{ECCV'24} & FSS   & ResNet50   & 40.6 & 93.9 & \ \ \underline{5.8} & 47.5 & 37.1 & 40.5 & 37.4 & 28.6 & 97.3 &\ \  2.8 & 27.7 & 32.5 & 27.2 & 27.1 \\
            DANet~\cite{fsvos}      & \pub{CVPR'21} & FSVOS & ResNet50   & \underline{45.4} & 97.1 & \ \ 8.2 & 41.4 & 44.7 & \underline{47.1} & \underline{48.2} & 25.4 & 77.2 & \underline{28.0} & 27.4 & 23.5 & 25.8 & 25.0 \\
            HPAN~\cite{hpan}        & \pub{CSVT'24} & FSVOS & ResNet50   & 44.4 & 97.3 & \ \ 7.2 & \underline{48.4} & \underline{45.2} & 43.4 & 40.8 & 34.0 & 99.1 &\ \  3.1 & \underline{37.6} & 34.8 & \underline{34.6} & 29.1 \\
            TTI~\cite{tti}          & \pub{IJCV'25} & FSVOS & ResNet50   & 45.2 & 97.6 &\ \  \underline{9.4} & 45.8 & 43.9 & 43.7 & 47.4 & \underline{35.6} & 70.6 & 26.2 & 33.8 & \underline{35.9} & 34.8 & \underline{37.8} \\
            \rowcolor{tablecolor} \textbf{DMA (Ours)} & \pub{ICCV'25}    & FSVOS & ResNet50   & \textbf{50.1} & \textbf{98.6} & \textbf{11.5} & \textbf{51.2} & \textbf{46.2} & \textbf{54.3} & \textbf{48.6} & \textbf{40.2} & \textbf{99.5} & \textbf{28.7} & \textbf{40.7} & \textbf{38.9} & \textbf{41.3} & \textbf{39.7} \\
    \mymidrule
            DANet~\cite{fsvos}      & \pub{CVPR'21} & FSVOS & VSwin-T    & \underline{49.8} & \underline{93.4} & \underline{16.5} & \underline{49.3} & \underline{47.5} & \underline{52.5} & \underline{49.9} & \underline{36.1} & \underline{37.2} & \underline{30.3} & \underline{34.8} & \underline{34.3} & \underline{38.3} & \underline{37.1} \\
            \rowcolor{tablecolor} \textbf{DMA (Ours)} & \pub{ICCV'25}       & FSVOS & VSwin-T    & \textbf{51.5} & \textbf{98.9} & \textbf{21.2} & \textbf{51.1} & \textbf{48.6} & \textbf{56.3} & \textbf{50.0} & \textbf{41.4} & \textbf{99.8} & \textbf{31.0} & \textbf{41.5} & \textbf{39.8} & \textbf{42.7} & \textbf{41.1} \\
    \myrule
    \end{tabular}
\end{table*}

\begin{table*}[t!]
    \footnotesize
    \centering
    \setlength\tabcolsep{3.4pt}
    \caption{Main results on \ourdataset benchmark with non-overlapping split (NS) setting. }
    \vspace{-3mm}
    \label{tab:MOVEOS}
    \begin{tabular}{lccccccccccccccccc}
    \myrule
    \multirow{2}{*}{\centering Methods} &
      \multirow{2}{*}{\centering Venue} &
      \multirow{2}{*}{\centering Type} &
      \multirow{2}{*}{\centering Backbone} &
      \multicolumn{3}{c}{Mean (2-way-1-shot)} &
      \multicolumn{4}{c}{\metricjf (2-way-1-shot)} &
      \multicolumn{3}{c}{Mean (5-way-1-shot)} &
      \multicolumn{4}{c}{\metricjf (5-way-1-shot)} \\
      \cmidrule(lr){5-7} \cmidrule(lr){8-11} \cmidrule(lr){12-14} \cmidrule(lr){15-18}
                       &         &       &             & \metricjf & T-Acc & N-Acc & 1    & 2    & 3    & 4    & \metricjf & T-Acc & N-Acc & 1    & 2    & 3     & 4    \\
      \midrule      
            LMPM~\cite{mevis}       & \pub{ICCV'23} & RVOS  & ResNet50   & 38.8 & 94.8 & \ \ 4.4 & 45.5 & 34.5 & 36.5 & 38.5 & 29.8 & 96.6 & \ \ 2.4 & 28.9 & 26.4 & \underline{37.6} & 26.1 \\
            CyCTR~\cite{cyctr}      & \pub{ECCV'24} & FSS   & ResNet50   & 28.2 & \underline{98.0} & \ \ 1.0 & 31.2 & 25.7 & 33.0 & 22.7 & 23.4 & 95.3 &\ \  3.2 & 23.6 & 21.2 & 27.6 & 21.3 \\
            SCCAN~\cite{sccan}      & \pub{ECCV'24} & FSS   & ResNet50   & 34.5 & 92.3 & \ \ \underline{5.9} & 41.8 & 32.7 & 29.2 & 34.3 & 27.8 & 96.0 & \ \ \underline{4.3} & 31.7 & \underline{31.4} & 25.8 & 22.2 \\
            DANet~\cite{fsvos}      & \pub{CVPR'21} & FSVOS & ResNet50   & \underline{44.6} & 97.7 &\ \  2.5 & 48.4 & \underline{36.9} & 49.5 & 43.6 & 29.9 & 96.6 &\ \  4.2 & 29.5 & 24.0 & 31.0 & \underline{35.3} \\
            HPAN~\cite{hpan}        & \pub{CSVT'24} & FSVOS & ResNet50   & 39.1 & 96.3 &\ \  1.4 & \underline{49.1} & 34.9 & 40.0 & 32.3 & 30.2 & \underline{99.1} & \ \ 1.1 & \underline{35.3} & 28.5 & 30.3 & 26.6 \\
            TTI~\cite{tti}          & \pub{IJCV'25} & FSVOS & ResNet50   & 43.6 & 97.2 &\ \  2.4 & 47.2 & 33.4 & \underline{50.0} & \underline{43.9} & \underline{32.7} & 98.3 &\ \  0.9 & 35.0 & 29.8 & 35.7 & 30.2 \\
            \rowcolor{tablecolor} \textbf{DMA (Ours)} & \pub{ICCV'25}    & FSVOS & ResNet50   & \textbf{46.0} & \textbf{98.2} & \ \ \textbf{7.8} & \textbf{47.8} & \textbf{37.9} & \textbf{48.0} & \textbf{50.3} & \textbf{34.7} & \textbf{99.6} & \ \ \textbf{5.0} & \textbf{35.6} & \textbf{31.5} & \textbf{37.0} & \textbf{34.6} \\
    \mymidrule
            DANet~\cite{fsvos}      & \pub{CVPR'21} & FSVOS & VSwin-T    & \underline{47.4} & \underline{97.2} & \ \ \underline{1.2} & \underline{53.2} & \underline{37.4} & \underline{48.3} & \underline{50.9} & \underline{30.0} & \underline{74.8} &\ \  \underline{2.9} & \underline{34.6} & \underline{26.5} & \underline{32.1} & \underline{26.8} \\
            \rowcolor{tablecolor} \textbf{DMA (Ours)} & \pub{ICCV'25}       & FSVOS & VSwin-T    & \textbf{49.0} & \textbf{98.0} & \textbf{8.8} & \textbf{54.4} & \textbf{37.4} & \textbf{48.5} & \textbf{55.9} & \textbf{35.4} & \textbf{97.4} & \textbf{9.3} & \textbf{37.9} & \textbf{29.9} & \textbf{36.6} & \textbf{37.2} \\
    \myrule
    \end{tabular}
    \vspace{-3mm}
\end{table*}

\subsection{Benchmark Necessity Study}
To demonstrate the necessity of our \ourdataset benchmark, we conduct experiments comparing state-of-the-art methods across different areas on both the common-used YouTube-VIS~(YTVIS)~\cite{YouTubeViS} and our proposed \ourdataset datasets, as shown in \Cref{tab:necessity_study}. The use of YouTube-VIS strictly follows the few-shot setting in \cite{fsvos}.
Notably, when evaluated on YTVIS, the image-based FSS method SCCAN~\cite{sccan} achieves 62.3\% \metricjf, comparable to HPAN~\cite{fsvos} (63.0\% \metricjf) which is specifically designed for FSVOS. This suggests that YTVIS primarily relies on category-based object association, requiring minimal temporal understanding between support and query samples.
However, the performance landscape changes dramatically on \ourdataset. SCCAN's performance drops significantly to 40.6\% \metricjf, substantially lower than HPAN's 44.4\% \metricjf. This stark contrast highlights the critical role of temporal information in \ourdataset. Furthermore, we enhance HPAN (denoted as HPAN*) by incorporating temporal modeling during prototype extraction through a simple self-attention mechanism across frames. This modification yields a notable improvement from 44.4\% to 46.3\% \metricjf, further emphasizing the importance of motion understanding in our benchmark.

Furthermore, we benchmark the referring video object segmentation method LMPM \cite{mevis} on our \ourdataset dataset by converting support set into referring expressions with the template ``\textit{The one [\textbf{motion category}]}''. 
While LMPM achieves competitive performance of 62.5\% \metricjf on YTVIS, only 0.2\% \metricjf lower than the temporally-enhanced HPAN*, its performance drops significantly to 41.8\% \metricjf on \ourdataset, substantially underperforming compared to HPAN*'s 46.3\% \metricjf. 
We attribute this performance gap to the presence of fine-grained, novel, and specialized motion patterns in our \ourdataset dataset like mutations and moonwalks, which are difficult to describe clearly using text. These findings underscore the unique challenges posed by \ourdataset and emphasize its necessity in advancing motion-centric few-shot video understanding.

\subsection{Main Results}
\label{sec:main_results}
\vspace{-1mm}
As shown in~\Cref{tab:MOVENS} and~\Cref{tab:MOVEOS}, we benchmark referring video object segmentation (RVOS) \cite{mevis}, few-shot image segmentation (FSS)~\cite{cyctr,sccan}, and few-shot video object segmentation (FSVOS) methods~\cite{fsvos,hpan,tti} across \textit{2-way-1-shot} and \textit{5-way-1-shot} test settings on both OS and NS data splits with two different backbones, ResNet50~\cite{resnet} and VideoSwin-T~\cite{videoswin}. Our proposed \ourmodel consistently outperforms all competing methods across all metrics and settings, demonstrating its superior few-shot motion understanding and segmentation capabilities. For the \metricjf metric with ResNet50 backbone, DMA achieves significant improvements over the second-best method, reaching 50.1\% (\vs 45.4\%) in \textit{2-way-1-shot} and 40.2\% (\vs 35.6\%) in \textit{5-way-1-shot} under the OS setting. This substantial performance gap highlights the limitations of existing methods in effectively modeling motion patterns. When using VideoSwin-T backbone, which provides better temporal feature extraction, our method further improves to 51.5\% and 41.4\% in respective settings, indicating the importance of temporal modeling in motion-centric segmentation.
It is worth noting that performance on the NS setting (46.0\% \metricjf with ResNet50) is lower than OS (50.8\%), reflecting its greater challenge as a more realistic scenario where test categories have completely different parent classes from training categories. Regarding robustness metrics, while DMA maintains high target accuracy (T-Acc) of 98.6\% and achieves better non-target accuracy (N-Acc) of 11.5\% with ResNet50 compared to baselines, the generally low N-Acc scores across all methods suggest a common challenge in effectively modeling background information to reduce false positives. This limitation points to a promising direction for future research in \ourdataset.

\subsection{Ablation Studies}
\label{sec:ablation}
\begin{table}
    \centering
    \small
    \setlength{\tabcolsep}{8pt}
    \caption{Ablation study on motion extractor.}
    \vspace{-3mm}
    \begin{tabular}{l|c|ccc} 
    \myrule
        ID & Motion Extractor & \multicolumn{1}{c}{\metricjf} & \multicolumn{1}{c}{T-acc} & \multicolumn{1}{c}{N-acc} \\
        \hline
        I   & Mask Pooling   &  41.3 &  98.0 &  \ \ 6.8 \\
        II  & Mask Adapter   &  43.4 &  98.4 &  \ \ 6.6 \\
        III & Differencing   & \textbf{46.8}  &  \textbf{99.8} &  \textbf{12.3} \\
    \myrule
    \end{tabular}
    \label{tab:ablation_study_1}
    \vspace{-1.6mm}
\end{table}
\begin{table}
    \centering
    \small
    \setlength{\tabcolsep}{8pt}
    \caption{Ablation study on DMA prototype extractor.}
    \vspace{-3mm}
    \begin{tabular}{l|cc|ccc} 
    \myrule
        ID & Appear. & Motion & \multicolumn{1}{c}{\metricjf} & \multicolumn{1}{c}{T-acc} & \multicolumn{1}{c}{N-acc} \\
        \hline
        I   & \cmark    & \xmarkg   &  36.5 &  80.1 &  \textbf{30.4} \\
        II  & \xmarkg   & \cmark    &  43.8 &  95.5 &  10.7  \\
        III & \cmark    & \cmark    &  \textbf{46.8} &  \textbf{99.8} &  12.3  \\
    \myrule
    \end{tabular}
    \label{tab:ablation_study_2}
    \vspace{-2.6mm}
\end{table}

\myparagraph{Ablation study on motion extractor.}
As shown in \Cref{tab:ablation_study_1}, our proposed differencing-based motion extractor achieves better performance compared to baseline approaches such as mask pooling and mask adapter~\cite{mask_adapter}, improving \metricjf from 41.3\% (mask pooling) and 43.4\% (mask adapter) to 46.8\%. The explicit motion modeling through frame differencing enables more effective extraction of motion-centric prototypes, leading to enhanced motion pattern recognition and segmentation performance.

\myparagraph{Ablation study on DMA prototype extractor.}
As shown in \Cref{tab:ablation_study_2}, we analyze the contribution of appearance and motion prototypes in our DMA prototype extractor. Using only appearance prototypes (I) achieves 36.5\% \metricjf, while using only motion prototypes (II) results in 43.8\% \metricjf. {These results suggest that while appearance features provide static cues for target object recognition, they are insufficient for motion-centric video understanding in \ourdataset. In contrast, motion features capture temporal dynamics, enhancing the distinction between different motion categories. When combining both prototypes (III), our model achieves the best performance of 46.8\% \metricjf, demonstrating the complementary nature of static appearance and dynamic motion information in our DMA mechanism.}

\myparagraph{Ablation study on auxiliary classification.}~As shown in \Cref{tab:ablation_category_head}, applying auxiliary classification supervision separately to appearance and motion prototypes yields the best performance of 46.8\% \metricjf. We attribute this improvement to explicit supervision of object and motion, which effectively enhances the decoupling of motion and appearance features, resulting in overall performance gains.

\begin{table}
    \centering
    \small
    \setlength{\tabcolsep}{8pt}
    \caption{Ablation study on auxiliary classification.}
    \vspace{-3mm}
    \begin{tabular}{l|cc|ccc} 
    \myrule
        ID  & Object    & Motion & \multicolumn{1}{c}{\metricjf} & \multicolumn{1}{c}{T-acc} & \multicolumn{1}{c}{N-acc} \\
        \hline
        I   & \xmarkg   & \xmarkg   &  43.8 &  97.2 &  \ \ 5.2 \\
        II  & \xmarkg   & \cmark    &  44.2 &  87.6 &  \ \ 9.6  \\
        III & \cmark    & \xmarkg   & 43.5  &  83.2 &  \ \ 7.2  \\
        IV  & \cmark    & \cmark    &  \textbf{46.8} &  \textbf{99.8} &  \textbf{12.3}  \\
    \myrule
    \end{tabular}
    \label{tab:ablation_category_head}
    \vspace{-1.6mm}
\end{table}
\begin{table}
    \centering
    \small
    \setlength{\tabcolsep}{8pt}
    \caption{Oracle results on motion category and mask.}
    \vspace{-3mm}
    \begin{tabular}{l|cc|ccc} 
    \myrule
        ID & Motion & Mask & \metricjf & T-acc & N-acc \\
        \hline
        I   & \cmark    & \xmarkg   &  63.6 &  100.0 &  100.0 \\
        II  & \xmarkg     & \cmark   &  74.3 &  \ \ 73.2 &  100.0 \\
    \myrule
    \end{tabular}
    \label{tab:oracle_results}
    \vspace{-1.6mm}
\end{table}

\myparagraph{Orcale results.}
As shown in \Cref{tab:oracle_results}, we conduct oracle experiments to analyze the performance upper bound of our model. When provided with perfect motion category labels, the model achieves 63.6\% \metricjf, while using ground truth masks yields higher \metricjf of 74.3\%. The results indicate significant room for improvement in both motion understanding and mask prediction capabilities.

\begin{figure}
    \centering
    \includegraphics[width=0.95\linewidth]{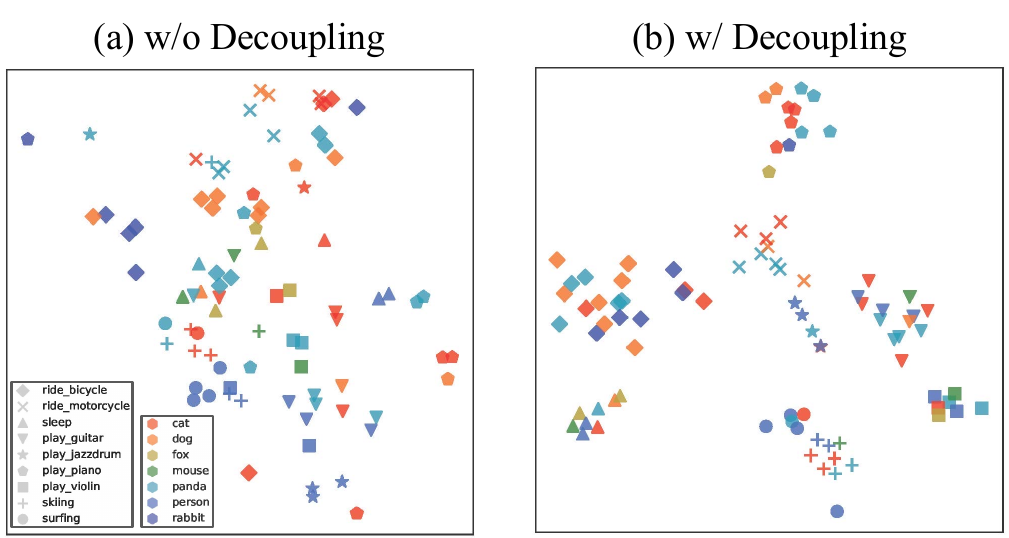}
    \vspace{-3mm}
    \caption{t-SNE~\cite{tsne} visualization of prototypes in our model (a) \textbf{w/o decoupling} and (b) \textbf{w/ decoupling}. Different colors and different shapes represent the object categories (\eg, cat) and motion categories (\eg, surfing), respectively. The proposed \ourmodel effectively extracts the motion-centric prototypes and makes those having the same motions closer in feature space.
    }
    \label{fig:vis_prototype}
    \vspace{-3mm}
\end{figure}

\begin{figure*}[t!]
    \centering
    \includegraphics[width=0.998\linewidth]{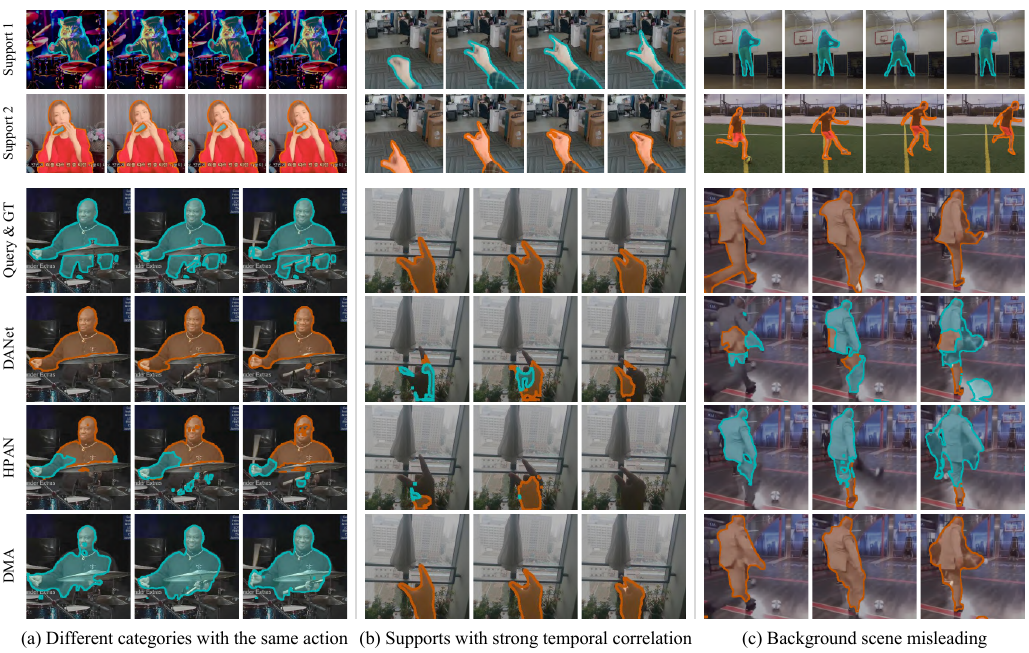}
    \vspace{-6mm}
    \caption{Qualitative comparison of representative cases from \ourdataset between baseline methods, DANet~\cite{fsvos} and HPAN~\cite{hpan}, and our proposed \ourmodel. (a) shows different object categories of ``\textit{cat}'' (Support 1) and ``\textit{person}'' (Query) performing the same action, ``\textit{playing drums}''. (b) presents temporally correlated motions: fingers transitioning ``\textit{from pinching to opening}'' (Support 1) and ``\textit{from opening to pinching}'' (Support 2 \& Query videos). (c) is a misleading background in the Query video, playing ``\textit{football}'' on the ``\textit{basketball court}''.}
    \label{fig:fig_qualitative}
    \vspace{-2.6mm}

\end{figure*}

\myparagraph{t-SNE Visualzation of Prototypes}.
As shown in \Cref{fig:vis_prototype}, we visualize the decoupled motion-appearance prototypes $P_{\text{dma}}$ using t-SNE~\cite{tsne}. Without our DMA approach for prototype extraction, prototypes cluster according to object categories, \ie, colors.~In contrast, with our proposed DMA approach, prototypes cluster based on motion categories, \ie, shapes, highlighting the effectiveness of our method in capturing motion-centric representations rather than appearance-based features.

\subsection{Qualitative Results}
Figure~\ref{fig:fig_qualitative} presents several representative examples comparing our \ourmodel with the baseline methods DANet~\cite{fsvos} and HPAN \cite{hpan}. 
In case (a), we showcase a challenging scenario where objects of different categories perform the same action: a cat playing the drums and a person playing the flute in the support videos while a person playing the drums in the query video. Baseline methods fail by segmenting based on the same object category of ``person'', whereas our method correctly segments the target based on the shared motion pattern, ``playing the drums''. This demonstrates the effectiveness of our DMA design in prioritizing motion cues over object class identity.
In case (b), we highlight a scenario with strong temporal correlations between frames in the support set: fingers transitioning from pinching to opening and from opening to pinching. While baseline methods struggle with fine-grained action discrimination due to insufficient temporal modeling, our proposed method effectively captures subtle temporal dependencies, leading to precise motion recognition and object segmentation.
In case (c), our model correctly segments the target object, whereas the baseline methods are misled by the background context of playing ``football'' on the ``basketball court'', and fail to capture the specific motion category. These examples demonstrate the superiority of our approach in few-shot motion understanding. Additional failure cases are provided in the supplementary material.

\section{Conclusion}
\label{sec:conclusion}
We introduce \ourdataset, a new benchmark for motion-guided few-shot video object segmentation.~Unlike existing FSVOS datasets that segment objects based on object categories, the proposed \ourdataset emphasizes temporal dynamics by segmenting objects according to motion categories that correlate with support and query videos.
Experimental results show that \ourdataset poses significant challenges to current state-of-the-art methods, motivating us to propose \ourmodel that decouples motion and appearance prototypes for more robust and effective motion prototype extraction. 
\ourdataset provides a foundation for advancing research in motion-centric few-shot video understanding and temporal modeling for segmentation tasks.

\myparagraph{Future Directions.}
The \ourdataset benchmark opens up several promising research directions that need further investigation.
We highlight some key areas for future exploration:
i) decomposing complex motions into meta-motions for more general and efficient motion prototype learning,
ii) modeling relational motions that involve interactions between multiple objects,
iii) improving fine-grained motion discrimination through extracting more robust motion prototypes,
iv) handling long-term temporal motions spanning multiple seconds through efficient temporal modeling, and 
v) learning discriminative background prototypes to suppress false positives in complex scenes better.

\footnotesize{\paragraph{Acknowledgement.}~This project was supported by the National Natural Science Foundation of China (NSFC) under Grant No. 62472104.}

{
    \small
    \bibliographystyle{ieeenat_fullname}
    \bibliography{main}
}

\end{document}